\documentclass[conference,a4paper]{APSIPA2025}
\usepackage{amsmath}
\usepackage{amsfonts}
\usepackage{graphicx}
\usepackage{multirow}
\usepackage{threeparttable}
\usepackage{booktabs}
\usepackage[backend=biber,style=ieee,citestyle=numeric-comp]{biblatex}

\usepackage{geometry}
\usepackage{fancyhdr}

\fancypagestyle{firststyle}{
  \fancyhf{}
  \fancyhead[C]{2025 Asia Pacific Signal and Information Processing Association Annual Summit and Conference (APSIPA ASC)}
}

\begin{document}

\title{Kernel Ridge Regression for Efficient Learning of High-Capacity Hopfield Networks}

\author{
\authorblockN{
Akira Tamamori
}

\authorblockA{
Aichi Institute of Technology, Japan}
}

\maketitle
\thispagestyle{firststyle}
\pagestyle{fancy}
\pagenumbering{gobble}          

\begin{abstract}
  Hopfield networks using Hebbian learning suffer from limited storage
  capacity. While supervised methods like Linear Logistic Regression
  (LLR) offer some improvement, kernel methods like Kernel Logistic
  Regression (KLR) significantly enhance storage capacity and noise
  robustness. However, KLR requires computationally expensive
  iterative learning. We propose Kernel Ridge Regression (KRR) as an
  efficient kernel-based alternative for learning high-capacity
  Hopfield networks. KRR utilizes the kernel trick and predicts
  bipolar states via regression, crucially offering a non-iterative,
  closed-form solution for learning dual variables. We evaluate KRR
  and compare its performance against Hebbian, LLR, and KLR. Our
  results demonstrate that KRR achieves state-of-the-art storage
  capacity (reaching a storage load of 1.5) and noise robustness,
  comparable to KLR. Crucially, KRR drastically reduces training time,
  being orders of magnitude faster than LLR and significantly faster
  than KLR, especially at higher storage loads. This establishes KRR
  as a potent and highly efficient method for building
  high-performance associative memories, providing comparable
  performance to KLR with substantial training speed advantages. This
  work provides the first empirical comparison between KRR and KLR in
  the context of Hopfield network learning.
\end{abstract}

\section{Introduction}

Hopfield networks~\cite{1} stand as a cornerstone in the study of
recurrent neural networks and associative memory. They provide a
simple yet powerful model for content-addressable memory, where stored
patterns correspond to attractors of the network dynamics. However,
the classical Hebbian learning rule~\cite{2} severely limits the
storage capacity $(\beta = P/N)$ to a pattern-to-neuron ratio of
$\beta \approx 0.14$, leading to the emergence of spurious states and
recall failure at higher loads.

While originally proposed in the 1980s, attractor-based memory models
have seen a resurgence of interest, partly due to their theoretical
connections to the attention mechanisms in modern
Transformers~\cite{3} and their potential for energy-efficient,
brain-inspired computing hardware. This renewed interest motivates the
development of learning algorithms that can overcome the classical
capacity limitations and create robust, high-capacity associative
memories.

To this end, research has shifted towards supervised learning
paradigms. Linear methods like Linear Logistic Regression
(LLR)~\cite{4} offer moderate improvements but are fundamentally
constrained by the linear separability of patterns. Non-linear
approaches, particularly those employing the kernel trick, provide a
more powerful alternative. By implicitly mapping patterns into
high-dimensional feature spaces, kernel methods can effectively handle
complex, non-linear relationships.

Kernel Logistic Regression (KLR), a well-established method in
statistical learning~\cite{5}, has recently been demonstrated as a
highly effective learning algorithm for Hopfield
networks~\cite{6}. Our previous work~\cite{6} showed that KLR
dramatically enhances both storage capacity and noise robustness,
achieving state-of-the-art recall performance. However, KLR relies on
iterative optimization, which can be computationally expensive and
time-consuming, particularly for large datasets, limiting its
practical scalability.

In this paper, we propose Kernel Ridge Regression (KRR)~\cite{7} as a
powerful and highly efficient alternative. KRR frames the learning
problem as a regression task and, crucially, offers a non-iterative,
closed-form solution for its parameters. This suggests a substantial
advantage in training speed. A key advantage of KRR is that the
optimal parameters (dual variables) can be obtained via a direct,
analytical closed-form solution, eliminating the need for
computationally expensive iterative optimization.

Our primary goal is to empirically investigate whether KRR can match
the state-of-the-art recall performance (capacity and noise
robustness) of KLR while offering significantly faster training. We
conduct a comprehensive experimental evaluation comparing KRR against
Hebbian learning, LLR, and KLR. This study presents, to the best of
our knowledge, the first direct empirical comparison between KRR and
KLR for training Hopfield networks, highlighting the trade-offs
between recall performance and learning efficiency. Our findings
establish KRR as a potent and practical method for building
high-performance associative memories.

\section{Methods}
\subsection{Model Setup}
We consider a network of \(N\) bipolar neurons \(\{-1, 1\}^{N}\). Let
\(\{\boldsymbol{\xi}^{\mu}\}_{\mu=1}^{P}\) be the set of \(P\) bipolar
patterns to be stored in the network's associative memory.  We define
a pattern matrix $\mathbf{X} \in \mathbb{R}^{P\times N}$ where the
$\mu$-th row is the pattern vector $(\boldsymbol{\xi}^{\mu})^{\top}$.
For supervised learning approaches requiring binary targets (LLR and
KLR), we transform these patterns into target vectors
\( \mathbf{t}^{\mu} \in \{0, 1\}^{N}\) where
\(t_i^{\mu} = (\xi_i^{\mu} + 1) / 2\). For KRR, the original bipolar
values \(\xi_{i}^{\mu}\) are used as targets. The network state at
discrete time \(t\) is represented by the vector
\(\mathbf{s}(t) \in \{-1, 1\}^{N}\).  This state evolves over
subsequent time steps according to the dynamics determined by the
specific learning algorithm employed (detailed in Sec. 2.2 and 2.3).

\subsection{Learning Algorithms}

We compare the proposed KRR method against several baseline and
related learning approaches.

\subsubsection{Baselines}

\begin{itemize}
\item \textbf{Hebbian Learning:}
  The standard Hebbian
  rule~\cite{1, 2} computes the weight matrix (diagonals zeroed) 
  $$ \mathbf{W}^{\text{Heb}} = \frac{1}{N} \mathbf{X}^{\top}\mathbf{X}.$$
   Recall uses \(s_i(t+1) = \text{sign}(\sum_{j\neq i} \mathbf{W}_{ij}^{\text{Heb}} \mathbf{s}_j(t))\)
  This serves as the classical baseline.
\item \textbf{Linear Logistic Regression (LLR):} A linear supervised
  baseline~\cite{4}, where weights
  \(\mathbf{W}^{\text{LLR}}\) are learned iteratively via gradient
  descent to optimize logistic loss with a linear predictor for
  predicting neuron states. Recall uses the same update rule as
  Hebbian, with \(\mathbf{W}^{\text{LLR}}\).
\item \textbf{Kernel Logistic Regression (KLR):} An iterative kernel-based
  method shown to achieve high storage capacity~\cite{6}. It learns
  dual variables \(\boldsymbol{\alpha}^{\text{KLR}}\) by iteratively optimizing
  regularized logistic loss. Recall involves kernel evaluations
  (Sec. 2.3). We use results from~\cite{6} for comparison.  
\end{itemize}

\subsubsection{Kernel Ridge Regression Learning}
As the primary focus of this work, we propose Kernel Ridge Regression
(KRR)~\cite{7} as an efficient supervised learning
approach. KRR frames the task as regression, aiming to directly
predict the target bipolar state \(\xi_{i}^{\mu} \in \{0, 1\}\) for
each neuron \(i\).

Leveraging the kernel trick with a chosen kernel function
\(K(\cdot, \cdot)\), KRR finds a predictor function of the form
\(f_{i}(\boldsymbol{\xi}^{\mu}) = \sum_{\mu=1}^{P}
K(\boldsymbol{\xi}^{\nu}, \boldsymbol{\xi}^{\mu}) \alpha_{\mu
  i}\). This function minimizes the sum of squared errors between the
predictions \(f_{i}(\boldsymbol{\xi}^{\mu})\) and the targets
\(\xi_{i}^{\nu}\) over all patterns \(\nu\), subject to L2
regularization on the function norm in the reproducing kernel Hilbert
space.

A key advantage of KRR is that the optimal dual variables \(\boldsymbol{\alpha}_i = ( \alpha_{1i}, \ldots, \alpha_{Pi})^{\top}\) for neuron \(i\)  can be obtained non-iteratively via the following closed-form solution:
\begin{equation}
\boldsymbol{\alpha}_i = (\mathbf{K} + \lambda \mathbf{I})^{-1} \mathbf{y}_{i}
\end{equation}
where \(\mathbf{K}\) is the \(P \times P\) kernel matrix (\(\mathbf{K}_{\nu \mu} = K(\boldsymbol{\xi}^{\nu}, \boldsymbol{\xi}^{\mu})\)), \(\mathbf{y}_{i} = (\xi_{i}^{1}, \ldots, \xi_{i}^{P})^{\top}\) 
is the target vector containing the bipolar states for neuron \(i\) 
across all patterns,  \(\mathbf{I}\) is the \(P\times P\) identity matrix,
\(\lambda > 0\) is the regularization parameter.

The entire \(P\times N\) dual variable matrix \(\boldsymbol{\alpha}\)
can thus be computed efficiently by solving the linear system
\((\mathbf{K} + \lambda \mathbf{I}) \boldsymbol{\alpha} =
\mathbf{X}\), where \(\mathbf{X}\) is the \(P\times N\) matrix of
stored patterns. This eliminates the need for iterative optimization
required by LLR and KLR.

\subsection{Recall Process}
The state update rule differs between the baseline/linear methods and
the kernel methods. After learning, the network state
\(\mathbf{s}(t) \in \{-1, 1\}^{N}\) evolves over discrete time steps~\(T\).

\begin{itemize}
\item \textbf{Hebbian and LLR:} \(s_{i}(t+1) = \text{sign}(\sum_{j\neq i} \mathbf{W}_{ij} s_{j}(t)) \)  (using \(\mathbf{W}^{\text{Heb}}\) or \(\mathbf{W}^{\text{LLR}}\)), where
  \(s_{i}(t)\) is the \(i\)-th element of the state \(\mathbf{s}(t)\).
\item \textbf{KLR and KRR:} Recall operates without an explicit
  \(N \times N\) weight matrix \(\mathbf{W}\), instead utilizing the
  learned \(P \times N\) dual variable matrix \(\boldsymbol{\alpha}\)
  and the stored patterns \(\mathbf{X}\). The update rule involves
  three steps, performed synchronously for all neurons:
\begin{enumerate}
\item Compute kernel values: \\\(\mathbf{k}_{\mathbf{s}(t)} = \left[ K(\mathbf{s}(t), \boldsymbol{\xi}^{1}) , \ldots, K(\mathbf{s}(t), \boldsymbol{\xi}^{P})\right]\) (size \(1 \times P\)).
\item Compute activation potential: \( \mathbf{h}(\mathbf{s(t)})= \mathbf{k}_{\mathbf{s}(t)}\boldsymbol{\alpha} \) (size \(1 \times N\)).
\item Update the state: Determine the next state using the sign
  function. We set the activation threshold \(\boldsymbol{\theta}\) to
  zero for simplicity: \(\mathbf{s}(t+1) = \text{sign}(\mathbf{h}(\mathbf{s(t)}))\) (size \(1 \times N\)).
\end{enumerate}
\end{itemize}

It is important to note that KLR/KRR recall involves kernel
computations between the current state and all \(P\) stored patterns
at each update step.

\section{Experiments}
\subsection{Experimental Setup}
All simulations were performed using networks composed of $N = 500$ bipolar neurons, where each neuron's state is either $+1$ or $-1$. This network size ($N=500$) was chosen to be large enough to observe the scaling behavior of different algorithms and small enough to allow for extensive simulations across varying parameters within a reasonable computational time frame on standard hardware. The patterns $\xi^{\mu} \in \{-1, 1\}^N$ to be stored were generated randomly, with each component $\xi^{\mu}_i$ chosen independently and uniformly from $\{-1, 1\}$. The probability $P(\xi^{\mu}_i = 1) = 0.5$. This choice of random, unbiased patterns represents a standard benchmark for evaluating the storage capacity of the Hopfield network.

We implemented and compared four distinct learning algorithms:
\begin{itemize}
\item \textbf{Hebbian:} The classical standard Hebbian rule~\cite{1, 2}, representing the traditional, non-supervised baseline. Its learning complexity is $O(N^2)$ for computing the weight matrix.
\item \textbf{LLR:} A linear supervised learning method~\cite{4}, serving as a stronger linear baseline that aims to make stored patterns stable fixed points. LLR learns linear classifiers for each neuron iteratively via gradient descent.
\item \textbf{KLR:} An advanced kernel-based supervised learning method~\cite{4}, known for its high capacity and robustness, implemented based on our previous work for direct comparison. KLR learns dual variables via iterative optimization.
\item \textbf{KRR:} Our proposed method (detailed in Section 2.2.2), a non-iterative, closed-form kernel-based supervised learning approach.
\end{itemize}

For both KLR and KRR, which utilize the kernel trick, we employed the Radial Basis Function (RBF) kernel, defined as $K(\mathbf{x}, \mathbf{y}) = \exp(-\gamma ||\mathbf{x} - \mathbf{y}||^2)$. The kernel width parameter $\gamma$ was set to $1/N$. This is a common heuristic that scales the kernel width inversely with the input dimension, aiming to capture relevant distances between patterns. An L2 regularization term was applied during learning for LLR, KLR, and KRR to prevent overfitting and improve generalization. The regularization parameter $\lambda$ was uniformly set to $0.01$ for these three methods. This value was selected based on preliminary experiments and findings reported in~\cite{6}, where it was shown to provide robust performance across various storage loads.

For the iterative learning methods, LLR and KLR, a fixed learning rate
$\eta = 0.1$ was used. The number of learning updates was set to 100
iterations for LLR and 200 iterations for KLR based on preliminary
experiments showing convergence. These numbers were chosen to ensure
sufficient convergence of the respective optimization processes based
on observations from our previous study~\cite{6} and preliminary runs,
while also providing a practical comparison point for learning
times. Using more iterations might slightly improve the performance of
LLR/KLR but would further increase their learning time disadvantage
relative to KRR.

Recall simulations were performed by initializing the network state
$\mathbf{s}(0)$ and updating it synchronously for $T = 25$ steps using
the dynamics defined by each learned model (Section 2.3). This number
of steps was found to be sufficient for the network state to converge
to a fixed point or a limit cycle in most cases. Recall success for a
stored pattern $\xi^\mu$ was defined as the final state
$\mathbf{s}(T)$ having a normalized overlap
$m(T) = \frac{1}{N} \sum_{i=1}^N s_i(T) \xi^\mu_i$ greater than
$0.95$.

All simulations were implemented in Python, utilizing standard libraries such as NumPy for numerical computations. Experiments were run on a standard desktop CPU (specifically, an Intel Core i9 processor).

\subsection{Storage Capacity and Noise Robustness Evaluation}
\begin{figure}[t]
\begin{center}
  \includegraphics[width=0.99\hsize]{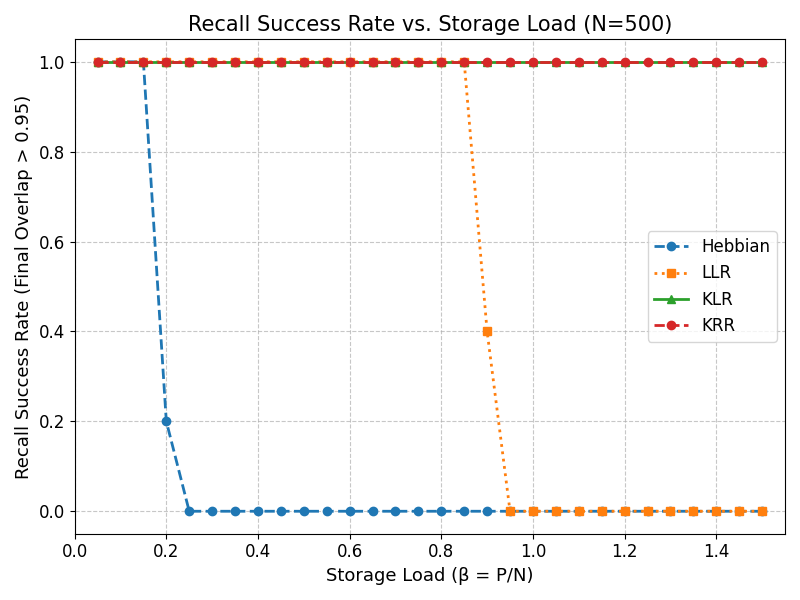}
   \caption{Recall success rate vs. storage load (\(\beta = P/N\)) for Hebbian, LLR, KLR and KRR (\(N=500\)). Note: The performance curve for KLR is virtually identical to KRR.}
\end{center}
\end{figure}

We first assessed the storage capacity of each learning algorithm by measuring the recall success rate as a function of the storage load $\beta = P/N$. For this evaluation, recall was initiated from a clean state corresponding to each stored pattern, i.e., $\mathbf{s}(0) = \xi^\mu$, testing the network's ability to maintain the learned patterns as stable fixed points. The storage load $\beta$ was varied from $0.05$ to $1.5$ in increments of $0.05$. For each value of $\beta$, $P = \lfloor \beta N \rfloor$ random patterns were generated and stored, and the recall success rate was computed as the percentage of these $P$ patterns that were successfully recalled according to our criterion ($m(T) > 0.95$).

Figure 1 plots the recall success rate against the storage load $\beta$ for $N = 500$ neurons. As anticipated from classical results, the standard Hebbian network exhibits a sharp phase transition and its performance drops significantly around the theoretical limit of $\beta \approx 0.14$~\cite{2}, quickly falling to 0\% success rate as $\beta$ increases further. LLR significantly improves upon this, maintaining a high, near-perfect recall rate up to approximately $\beta \approx 0.85$, but it fails completely as the load approaches $\beta = 0.95$ and beyond. In stark contrast, both KLR, consistent with our previous findings~\cite{6}, and the proposed KRR demonstrate substantially higher storage capacity. Both kernel methods achieved a remarkable $100\%$ recall success rate across the entire tested range of storage loads, including loads exceeding $\beta = 1.0$ (up to $\beta = 1.5$ shown in Figure 1). This empirical result strongly indicates their capability to reliably store and retrieve patterns even in highly overloaded regimes where $P > N$, a significant improvement over linear methods.

Next, we evaluated the noise robustness of the learned memories by examining the size of the basins of attraction. We fixed the storage load at an intermediate level, $\beta = 0.2$ (corresponding to $P = 100$ for $N = 500$ neurons), where both LLR and kernel methods typically succeed from clean initial states, but Hebbian learning already shows poor performance. For each stored pattern $\xi^\mu$, we generated 10 different corrupted initial states $\mathbf{s}(0)$ by flipping a certain percentage of bits relative to $\xi^\mu$ to achieve a desired initial overlap $m(0)$. The initial overlap $m(0) = \frac{1}{N} \sum_{i=1}^N s_i(0) \xi^\mu_i$ was varied from $0.0$ to $1.0$ in increments of $0.05$. We then measured the mean final overlap $m(T)$ achieved after $T=25$ recall steps, averaged over all $P=100$ stored patterns and the 10 trials per pattern.

Figure 2 shows the mean final overlap $m(T)$ as a function of the
initial overlap $m(0)$ at $\beta = 0.2$. As shown, the Hebbian network
fails to converge consistently to the target pattern regardless of the
initial overlap within this range, with mean final overlap remaining
low. LLR shows improved robustness, successfully recalling the pattern
(mean $m(T) \approx 1.0$) if the initial overlap $m(0)$ is above
approximately $0.4$. In contrast, both KLR and KRR exhibit
significantly enhanced robustness. They achieve perfect recall (mean
$m(T) \approx 1.0$) even when starting from much noisier states,
requiring an initial overlap $m(0)$ of only about $0.2$. This
demonstrates that both kernel methods create considerably larger
basins of attraction compared to LLR and, especially, the Hebbian
rule, making them more resilient to input noise and pattern
corruption. Notably, the robustness profiles of KLR and KRR appear
virtually identical under these conditions, reinforcing the
observation from the storage capacity test that their recall
performance characteristics are highly similar.

\begin{figure}[t]
\begin{center}
  \includegraphics[width=0.99\hsize]{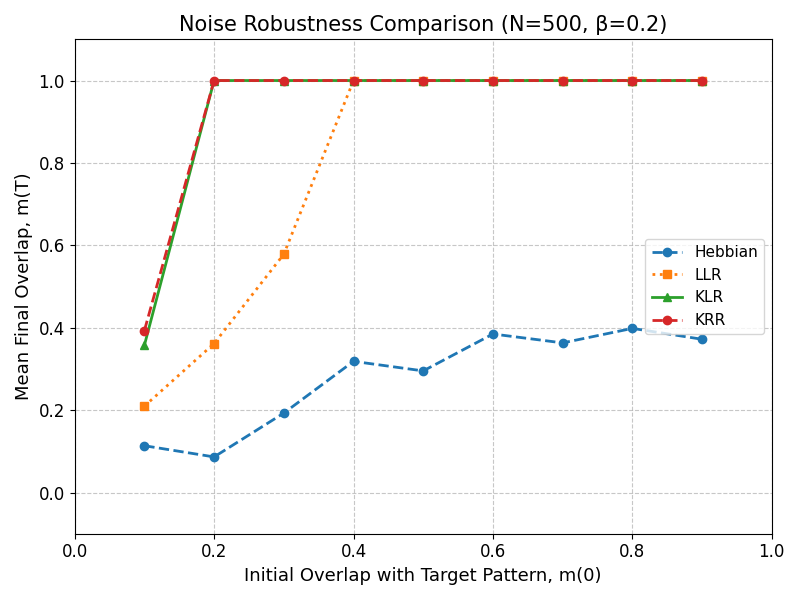}
  \caption{Noise robustness comparison: Final Overlap $m(T)$ vs. Initial Overlap $m(0)$ at $\beta=0.2$ for Hebbian, LLR, KLR and KRR. Note: The performance curve for KLR is virtually identical to KRR.}  
\end{center}
\end{figure}

\subsection{Learning Time Comparison}
To quantitatively assess the practical efficiency trade-offs alongside the performance benefits, we measured and compared the learning phase computational time for LLR, KLR, and KRR across various storage loads ($\beta$). The learning time was measured as the wall-clock time using Python's `\texttt{time.time()}' function, capturing the total duration required to complete the learning process for each method, including any setup and optimization steps. Measurements were performed on the standard desktop CPU mentioned in Section III.A. The results, averaged over 3 independent trials for each $\beta$ value, are presented in Table 1 and visualized in Figure 3. Note that the time axis in Figure 3 is on a logarithmic scale to effectively display differences spanning multiple orders of magnitude.

Hebbian learning, being non-iterative and computationally simple ($O(N^2)$ for weight matrix calculation), completes extremely quickly, typically in milliseconds for $N=500$. Its learning time is omitted from Table 1 and Figure 3 for clarity, as it is orders of magnitude faster than the supervised methods.

The timing results reveal significant differences in efficiency and scaling among the supervised methods. LLR, requiring 100 iterative updates of gradient descent on the network weights, exhibited the slowest performance. Its learning time increased approximately linearly with the storage load $\beta$ (and thus $P$), ranging from around 3 seconds at $\beta = 0.1$ to around 30 seconds at $\beta = 1.0$. This scaling is consistent with the need to iterate and process information related to $P$ patterns across $N$ neurons in each update.

KLR, using 200 iterative updates, was considerably faster than LLR across all tested loads, benefiting from operating in the dual space and potentially faster convergence properties for the logistic loss. Its learning time, however, still showed a clear dependence on $\beta$, increasing from around 0.07 seconds at $\beta = 0.1$ to around 1.35 seconds at $\beta = 1.0$. This increase is likely dominated by computations involving the $P \times P$ kernel matrix and matrix operations within the iterative optimization process, which scale with $P$.

Crucially, KRR demonstrated outstanding learning efficiency. Leveraging its non-iterative, closed-form solution for the dual variables (Eq. 1), KRR consistently achieved the fastest learning times by a large margin across all tested $\beta$ values. Its learning time remained under 0.1 seconds even at the maximum load of $\beta = 1.0$ (specifically, approximately 0.07 seconds as shown in Table 1), showing only a very modest increase with $\beta$. This time is primarily composed of the one-time construction of the $P \times P$ kernel matrix ($O(P^2 N)$ or $O(P^2)$ depending on implementation) and the subsequent solving of the $P \times P$ linear system ($O(P^3)$ or faster).

At $\beta = 1.0$, where KLR and KRR achieve peak capacity, KRR was approximately 20 times faster than KLR (with 200 updates) and over 400 times faster than LLR (with 100 updates). This dramatic speed advantage conferred by its closed-form solution is a key finding of this work, highlighting KRR's practical superiority for rapid model training compared to iterative supervised methods, particularly in the high-capacity regime. While the recall complexity $O(PN)$ remains a shared consideration for both KLR and KRR (Section 2.3), the learning time advantage of KRR is substantial.

\begin{table}[t]
  \centering
  \caption{Mean learning time (seconds) for different algorithms and storage loads (\(\beta = P/N\)) with \(N=500\). Values shown are mean \(\pm\) standard deviation over 3 trials.}
  \label{tab:timing_vs_alpha}
  \begin{tabular}{c c c c}
    \toprule
    Storage Load
    & \multicolumn{3}{c}{Learning Algorithm} \\
    \cmidrule(lr){2-4}
    $\beta$ & \shortstack{LLR \\ (100 Updates)} & \shortstack{KLR \\ (200 Updates)} & \shortstack{KRR \\ (Closed-form)} \\
    \midrule
    0.1 & $3.010 \pm 0.045$ & $0.072 \pm 0.014$ & $0.009 \pm 0.001$ \\ 
    0.2 & $5.988 \pm 0.128$ & $0.188 \pm 0.031$ & $0.010 \pm 0.003$ \\ 
    0.4 & $12.180 \pm 0.090$ & $0.365 \pm 0.009$ & $0.026 \pm 0.003$ \\ 
    0.6 & $18.356 \pm 0.426$ & $0.723 \pm 0.042$ & $0.035 \pm 0.005$ \\ 
    0.8 & $24.350 \pm 0.310$ & $0.804 \pm 0.010$ & $0.066 \pm 0.017$ \\ 
    1.0 & $30.363 \pm 0.844$ & $1.350 \pm 0.088$ & $0.069 \pm 0.003$ \\ 
    \bottomrule
  \end{tabular}
\end{table}

\begin{figure}[t]
\begin{center}
  \includegraphics[width=0.99\hsize]{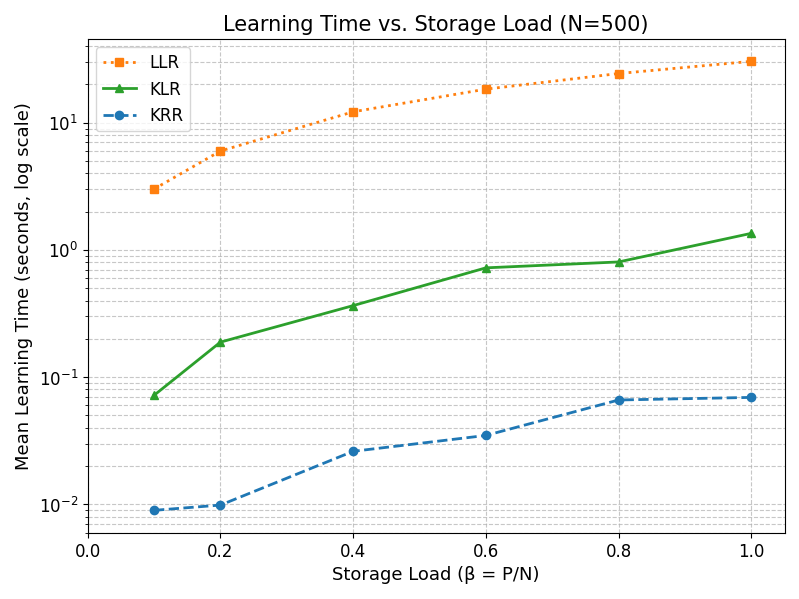}
  \caption{Learning time comparison (log scale) vs. storage load (\(\beta\)) for LLR, KLR, and KRR (\(N=500\), avg. over 3 trials).}
\end{center}
\end{figure}

\section{Discussion}
Our comprehensive experiments unequivocally demonstrate that
kernel-based supervised learning, using both KLR and KRR, dramatically
enhances the performance of Hopfield networks. These methods
substantially improve storage capacity and noise robustness compared
to the classical Hebbian rule and linear methods like LLR, pushing the
boundaries of what is achievable with these associative memory models.

\subsection{On the High Performance of Kernel Associative Memories}
The gains in storage capacity and noise robustness are particularly
remarkable. Both KLR and KRR achieved nearly perfect recall up to a
storage load of $\beta = 1.5$, far exceeding the Hebbian limit
($\beta \approx 0.14$) and LLR's practical limits
($\beta \approx 0.9$). This high performance, especially in the
overloaded regime ($P > N$), highlights their powerful
representational capability.

The superior performance of these kernel methods stems from the
"kernel trick"~\cite{8}, which implicitly maps input patterns into a
high-dimensional feature space. The Radial Basis Function (RBF)
kernel, used in our experiments, effectively transforms the pattern
space such that patterns that are non-linearly separable in the
original N-dimensional space become separable in this richer,
potentially infinite-dimensional feature space. This enhanced
separability allows the learning algorithms (KLR and KRR) to establish
more robust decision boundaries for each neuron, which translates into
stable fixed points with large basins of attraction for the stored
patterns. This mechanism effectively overcomes the primary
limitation of linear methods, which struggle when patterns are densely
packed and not linearly separable.


Notably, the recall performance profiles of KLR and KRR were virtually
identical in both capacity and noise robustness tests, as seen in
Figures 1 and 2. This is a significant finding, suggesting that for
the specific task of recalling bipolar patterns in a Hopfield-like
network, the choice between a logistic loss (KLR) and a squared-error
loss (KRR) is not critical to the final recall performance.

We hypothesize that this similarity in performance arises from the
interplay between the learning objectives and the nature of the recall
dynamics. The recall process relies on the sign of the activation
potential, $s_i(t+1) = \text{sign}(h_i(s(t)))$, effectively creating a
decision boundary at $h_i=0$. Although KLR and KRR optimize different
loss functions, both methods aim to map patterns to values with the
correct sign in the high-dimensional feature space. KLR, using
logistic loss, is designed to maximize the margin of separation for
classification, while KRR, using squared-error loss, aims to regress
the output towards the target values of $+1$ or $-1$. For bipolar targets,
both loss functions strongly penalize predictions that fall on the
wrong side of the zero threshold. Consequently, when combined with the
powerful non-linear mapping of the RBF kernel, both learning
algorithms are driven to find solutions that achieve a similar,
effective separation of patterns in the feature space. This likely
results in the formation of attractor landscapes with comparable fixed
point stability and basin of attraction sizes, leading to the observed
identical recall performance.

\subsection{Learning Efficiency: The Key Advantage of KRR}

While both kernel methods deliver comparable state-of-the-art recall
performance, the most significant practical difference, and a central
finding of this work, emerged in their learning efficiency. KRR's
non-iterative, closed-form solution for the dual variables (Eq. 1)
provides a dramatic speed advantage across all tested storage loads,
as shown in Table~1 and Figure~3.

At a storage load of $\beta = 1.0$, where both KLR and KRR achieve
peak capacity in our tests, KRR learned approximately 20 times faster
than our KLR implementation (with 200 updates) and over 400 times
faster than LLR (with 100 updates). While the iterative learning of
LLR and KLR scales with the number of patterns $P$ and the number of
iterations, KRR's learning time is dominated by the one-time
construction of the $P\times P$ kernel matrix and the subsequent
solving of a $P\times P$ linear system. This results in a much more
favorable scaling of learning time with respect to $P$, especially in
the high-capacity regime where $P$ is large.

This dramatic speed advantage conferred by its closed-form solution is
a key contribution of this work, highlighting KRR's practical
superiority for rapid model training. This efficiency makes KRR a
highly attractive and potent method for scenarios requiring frequent
retraining on large datasets or for applications where computational
resources for training are limited, establishing it as a highly
practical method for building high-performance associative memories.

\subsection{Scalability Considerations and Future Directions}
Despite the learning efficiency of KRR, the recall complexity for both
KLR and KRR, scaling as $O(PN)$, remains a key consideration. This
presents a scalability challenge for extremely large P, motivating
future research into kernel approximation techniques like the Nyström
method~\cite{9} or random features.

The significance of developing such high-capacity and efficient
associative memories extends beyond classical models. Recent
advancements have revealed deep connections between Hopfield networks
and the attention mechanisms in modern Transformers~\cite{3},
reigniting interest in attractor-based models. Efficiently trainable,
high-capacity associative memories could play a role in developing
more robust, interpretable, and computationally efficient AI systems.
Our work also aligns with and provides empirical support for the
broader framework of ``Kernel Memory Networks''~\cite{10}, which seeks to
unify various kernel-based approaches to memory modeling.

Based on our findings, several promising future directions emerge. A
detailed theoretical analysis of the attractor landscape, perhaps
using tools from statistical mechanics or dynamical systems theory,
could provide deeper insights into why these kernel methods work so
well. Furthermore, while this study focused on random bipolar patterns
as a standard benchmark, evaluating the performance of KRR on
real-world datasets, such as image or text data from established
benchmarks like MNIST or CIFAR-10, is a crucial next step to fully
assess its practical utility for tasks like pattern completion and
denoising. Developing online or incremental learning variants of KRR
and exploring other kernel types would also be valuable extensions.

\section{Conclusion}
We have demonstrated that Kernel Ridge Regression (KRR) provides a
powerful and highly efficient learning mechanism for Hopfield-type
associative memory networks. By leveraging kernel methods to
implicitly perform non-linear feature mapping, KRR achieves
exceptional storage capacity (successfully operating even for
\(P > N\)) and noise robustness, comparable to state-of-the-art
iterative methods like KLR. Critically, KRR achieves this high
performance with significantly reduced learning time due to its
non-iterative, closed-form solution, offering speedups of over an
order of magnitude compared to KLR and several orders of magnitude
compared to LLR in our tests. While the recall complexity scaling with
\(P\) remains a consideration for both KRR and KLR, KRR's combination
of top-tier storage performance and superior training efficiency makes
it a particularly compelling and practical approach for building
high-performance associative memory systems, especially when rapid
model development or retraining is required. This work further
highlights the potential of applying modern kernel-based machine
learning techniques to enhance and understand classic neural network
models.

\printbibliography

\end{document}